\PassOptionsToPackage{dvipsnames}{xcolor}

\documentclass{article}
\usepackage[utf8]{inputenc} 
\usepackage[T1]{fontenc}    
\usepackage{hyperref}       
\usepackage{url}            
\usepackage{booktabs}       
\usepackage{amsfonts}       
\usepackage{nicefrac}       
\usepackage{microtype}      
\usepackage{threeparttable}
\usepackage{wrapfig}
\usepackage{natbib}
\usepackage{xcolor}
\usepackage{amsmath}
\usepackage{xcolor}
\usepackage{arxiv}
\usepackage{url}
\newcommand\myshade{85}
\colorlet{mylinkcolor}{black}
\colorlet{mycitecolor}{violet}
\colorlet{myurlcolor}{YellowOrange}

\hypersetup{
  linkcolor  = mylinkcolor!\myshade!black,
  citecolor  = mycitecolor!\myshade!black,
  urlcolor   = myurlcolor!\myshade!black,
  colorlinks = true,
}

\usepackage{color}
\usepackage{times}
\usepackage{latexsym}
\usepackage[T1]{fontenc}
\usepackage[utf8]{inputenc}
\usepackage{microtype}
\usepackage{booktabs}
\usepackage{graphicx}
\usepackage{multirow}
\usepackage{float}
\usepackage[frozencache, cachedir=minted-cache]{minted}
\definecolor{navyblue}{RGB}{0,0,128}
\usepackage{bbding}
\usepackage{colortbl}
\usepackage{cleveref}
\usepackage{amsthm}
\usepackage{amssymb}

\title{
On Exploring PDE Modeling for Point Cloud Video Representation Learning
}

\author{
    \textbf{Zhuoxu Huang}\textsuperscript{1}\quad 
    \textbf{Zhenkun Fan}\textsuperscript{1}\quad 
    \textbf{Tao Xu}\textsuperscript{3}\quad
    \textbf{Jungong Han}\textsuperscript{1,2} \textsuperscript{\Envelope}\quad
\\
\small
    \textsuperscript{1}Aberystwyth University\quad \textsuperscript{2}University of Sheffield \quad 
\small \\
    \textsuperscript{3}Shanghai Investigation Design and Research Institute Co., Ltd. \quad
\\
\small
\texttt{zhh6@aber.ac.uk} \quad 
\texttt{} \quad \\
\texttt{\Envelope} \ corresponding author \texttt{jungonghan77@gmail.com}
}

\begin{document}

\maketitle

\begin{abstract}
Point cloud video representation learning is challenging due to complex structures and unordered spatial arrangement. Traditional methods struggle with frame-to-frame correlations and point-wise correspondence tracking. Recently, partial differential equations (PDE) have provided a new perspective in uniformly solving spatial-temporal data information within certain constraints. While tracking tangible point correspondence remains challenging, we propose to formalize point cloud video representation learning as a PDE-solving problem. Inspired by fluid analysis, where PDEs are used to solve the deformation of spatial shape over time, we employ PDE to solve the variations of spatial points affected by temporal information. By modeling spatial-temporal correlations, we aim to regularize spatial variations with temporal features, thereby enhancing representation learning in point cloud videos. We introduce Motion PointNet composed of a PointNet-like encoder and a PDE-solving module. Initially, we construct a lightweight yet effective encoder to model an initial state of the spatial variations. Subsequently, we develop our PDE-solving module in a parameterized latent space, tailored to address the spatio-temporal correlations inherent in point cloud video. The process of solving PDE is guided and refined by a contrastive learning structure, which is pivotal in reshaping the feature distribution, thereby optimizing the feature representation within point cloud video data. Remarkably, our Motion PointNet achieves an impressive accuracy of 97.52\% on the MSRAction-3D dataset, surpassing the current state-of-the-art in all aspects while consuming minimal resources (only 0.72M parameters and 0.82G FLOPs).
\end{abstract}

\section{Introduction}
\label{sec:intro}

Point cloud video modeling has emerged as a forefront topic in fields such as autonomous driving and robotics, primarily due to its ability to represent the variations in our 3D real world. A core problem in point cloud video modeling lies in efficient spatial-temporal representation learning \citep{song2022pref} from the intricate point cloud video data. 

\begin{wrapfigure}{r}{0.5\linewidth}
\vspace{-.3cm}
\begin{center}
\includegraphics[width=\linewidth]{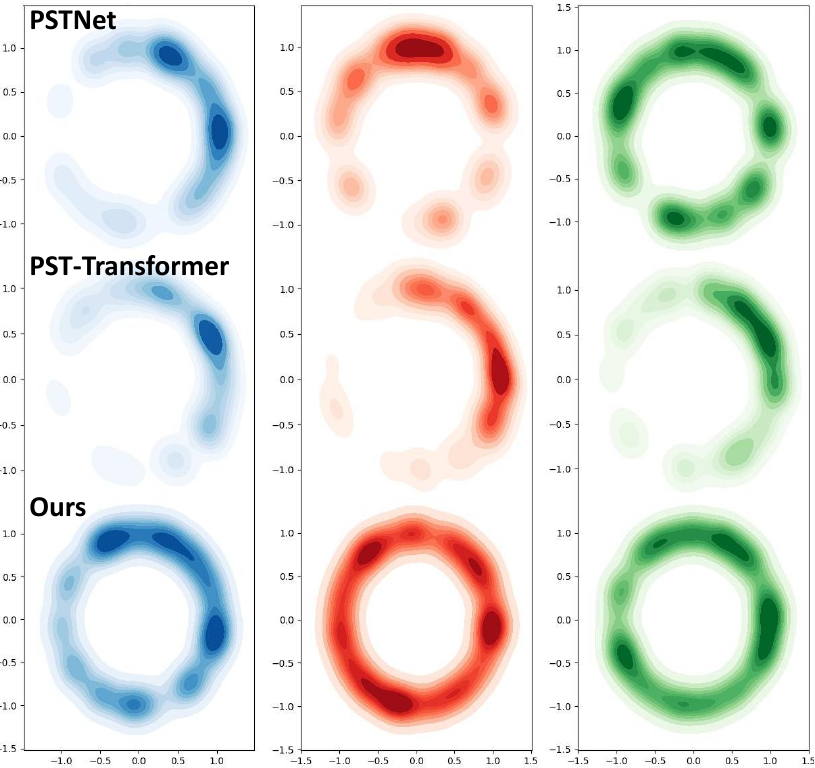}
\end{center}
\caption{Representations of MSRAction-3D test set on hypersphere. The temporal uniformity, spatial uniformity, and final logits uniformity are present in \textcolor{blue}{blue}, \textcolor{red}{red}, and \textcolor{green}{green}, respectively. Feature vectors should ideally be uniformly distributed over a unit hypersphere. The uniformity demonstrates the integrity of the information in features.}
\label{fig:uniformity}
\vspace{-.3cm}
\end{wrapfigure}

Unlike the continuous grid structure in traditional video data, point cloud videos are challenging to process. This is due to the point cloud video combining the unordered structure in spatial space and the ordered structure in temporal space. The uncertainty of correlation between frames hampers the application of explicit motion field capture in point cloud video learning. Without explicitly solving the point-wise correspondence, some standard models have been designed to extract dynamic information for point cloud video modeling. For instance, some methods adopt two-stream structures \citep{Zhong2022kinet, liu2021GeometryMotionnet, liu2022GeometryMotionTransformer} to decouple the spatial and temporal information for separate processing. Additionally, a popular model family \citep{fan2021pstnet, fan2022pstnet2, fan21p4transformer, fan2023psttransformer} employs both the spatial convolution and the temporal convolution with a sophisticated designed 4D convolution to directly process the point cloud video. Several fusion strategies \citep{liu2021GeometryMotionnet, liu2022GeometryMotionTransformer} and transformer-based post-process module \citep{fan21p4transformer, fan2023psttransformer} have been adopted after the process of spatial and temporal dimensions separately. However, the absence of motion field solving brings defects to these methods, where the spatial irregularity negatively impacts the temporal modeling, and vice versa. Our observation demonstrates that this drawback constrains the alignment and uniformity (see \Cref{fig:uniformity}) of feature distributions. Such alignment and uniformity have been demonstrated to exert beneficial effects on downstream tasks \citep{wang2020understanding}, especially for video learning \citep{shi2022Representation}. The mere fusion of disparate spatial and temporal features cannot effectively mitigate this issue, emphasizing the pressing need for a more effective method of learning point cloud video representations. We provide more details in the following \Cref{sec:preliminary}.

In this paper, we aim to improve the representation of point cloud video learning by effectively modeling the spatial-temporal correlations in point cloud video to enhance alignment and uniformity. To this end, we leverage partial differential equations (PDE), widely used in physical modeling, to formulate point cloud video representation learning as solvable PDEs. We demonstrate its effectiveness as an optimal method for spatio-temporal modeling within the point cloud video domain.

Our idea is inspired by the analogy between fluid systems in the physical domain and point cloud video representations. Firstly, both systems aim to study the deformation of spatial points over time due to motion, conceptualized as a unified representation across spatial and temporal dimensions. This representation is akin to the velocity field in fluid systems and the motion field in point cloud videos. PDEs are widely used as a powerful modeling tool to describe such phenomena involving changes in spatial and temporal dimensions \citep{tran2021factorized,liu2023htnet}. Secondly, PDE models applied to such physical systems are typically employed in discretized, high-dimensional coordinate spaces (\textit{i.e.} point cloud coordinate) to ensure solvability \citep{wu2023LSM}. These considerations make the application of PDEs in the point cloud video domain particularly intuitive. 

Our Motion PointNet is a pioneering approach that applies PDE-solving to point cloud video data. Specifically, we adopt PDE to model the spatial-temporal correlations and regularize the spatial variations with temporal features. It consists of two main components: Firstly, a \textbf{PointNet-like encoder} is adopted for the initial state of the spatial variations of points. Instead of processing the spatial and temporal dimensions separately, we adopted the set abstraction \citep{qi2017pointnet++} on the adjacent point cloud frames to generate local variations between frames. Secondly, we design our \textbf{PDE-solving module} to function as a cross-dimension (from temporal to spatial) feature reconstruction process from the masked token. The idea is to represent how spatial information is affected by temporal variations. We design a contrastive learning structure for this PDE-solving process. Our PDE-solving module functions by supervising the backbone through the backward propagation of the contrastive loss. This harmonizes the dichotomy between the comparative learning task and the downstream task \citep{khosla2020supervised}.

We conduct extensive experiments to demonstrate the significant advancements achieved by our Motion PointNet in the study of point cloud video. Our Motion PointNet improves the performance of the point cloud video action recognition task by a clear margin. Extensive experiments on various benchmarks including MSRAction-3D \citep{li2010msraction}, NTU RGB+D \citep{shahroudy2016ntu}, and UTD-MHAD \citep{utddataset} prove the superiority of our proposal. Prominently, with only 0.72M parameters and 0.82G FLOPs, our Motion PointNet achieves an accuracy of 97.52\% on the MSRAction-3D dataset. Furthermore, by leveraging the unsupervised contrastive learning mechanism, we explore the self-supervised potential of the PDE-solving module, offering pre-trained models for further investigation.

In summary, the contributions of this work can be outlined as follows:
\begin{itemize}
\item We propose a brand-new perspective that views the process of point cloud video representation learning as a PDE-solving problem. The novelty lies in the reconstruction of spatial information from the temporal dimension as the PDE-solving target. By doing so, we can establish a synthesis of unified representation across spatial spaces and temporal spaces, thereby enhancing the point cloud video representation learning.
\item We design a lightweight PointNet-like encoder and a PDE-solving module to capture the spatial-temporal correlations in the point cloud video. These components form the foundation of our Motion PointNet framework, tailored for the task of point cloud video understanding.
\end{itemize}

\section{Priori Observations}
\label{sec:preliminary}

To our knowledge, we are the first to adopt PDE models in the point cloud video domain. In this section, we explore the motivation behind this idea and verify the interpretability of point cloud video feature representation. We conduct proof-of-concept validation to reveal the intrinsic relationship between spatio-temporal representations and model performance.

\subsection{Preliminary}
While we require both spatial and temporal information to well present the point cloud video data, the lack of explicit motion field capturing in the current models could bring drawbacks in feature representation. Due to the inherent complexity of the spatial-temporal structure of point cloud video data, the mutual negative impacts between spatial and temporal modeling are inevitable \citep{fan2023psttransformer}. We quantify such drawbacks in the representation with the spatial-temporal alignment loss and the uniformity loss:

\begin{gather}
    \mathcal{L}_{\text{align}} \triangleq \left(\frac{1}{N}\sum_{i=1}^{N} \|x^S_i - x^T_i\|_2^\alpha\right) , ~ and ~ 
    \mathcal{L}_{\text{uniform}} \triangleq \log \left(\frac{1}{\binom{N}{2}} \sum_{i < j} e^{-t \| x_i - x_j \|_2^2} \right),
\end{gather}

where the $N$ represents the sample number of the validation data, $x$ represents a feature representation of a data sample, and superscript $S$ and $T$ stand for `spatial' and `temporal', respectively.

The alignment loss quantifies the discrepancy between a data sample's spatial and temporal dimensions. Ideally, both the spatial and temporal features can obtain the same logit distribution. Such alignment ensures consistent distribution between the spatial and temporal representation. Meanwhile, the uniformity loss indicates the thoroughness of the representation \citep{wang2020understanding}. Ideally, feature vectors should be uniformly distributed over a unit hypersphere to maintain the integrity of the information. We set $\alpha = 2$ and $t=2$ following the standard setting in \cite{wang2020understanding}.

\subsection{Alignment and Uniformity of Spatial-Temporal Representations}

We evaluate the alignment and uniformity of the representation from previous state-of-the-are models, PSTNet \citep{fan2021pstnet} and PST-Transformer \citep{fan2023psttransformer}, and compare them with our proposed method. The spatial and temporal representations are decomposed from the last-layer backbone feature, utilizing a pooling operation along the corresponding dimension. We perform the alignment loss and the uniformity loss in \Cref{tab:alignmentuniformity} and \Cref{fig:uniformity}, and show their effect on model performance.

PSTNet adopts a 4D convolution for point cloud video data, where the spatial and temporal information are separately abstracted from different convolutions. It maintains the uniformity of the representations to a certain extent. However, the performance of the model is hindered by the negative effects arising from processing the dimensions separately. To mitigate this issue, PST-Transformer adopts a transformer module after the 4D convolution layers. Although merging spatial and temporal features reduces the spatial-temporal alignment loss to some extent, it also disrupts the uniformity of the overall feature representations. Unlike others, our Motion PointNet preserves feature uniformity and enhances the alignment of spatial and temporal representations, thereby significantly boosting model performance. By adopting PDE-solving for unified spatio-temporal representation, our model outperforms the previous state-of-the-art to a great extent.

\begin{table}[htbp]
\caption{Alignment loss and uniformity loss of spatial-temporal representations from MSRAction-3D test set.}
\label{tab:alignmentuniformity}
\begin{center}
{
\begin{tabular}{lccccc}
  \toprule
  & alignment($\downarrow$) & \multicolumn{3}{c}{uniformity($\downarrow$)} & \multirow{2}{*}{Acc.(\%)($\uparrow$)} \\
  & S-T & T & S & logit & \\
  \midrule
  PSTNet \citep{fan2021pstnet} & 1.1421 & \underline{-1.2254} & \underline{-1.2501} & \underline{-1.5494} & 91.20\\
  PST-Transformer \citep{fan2023psttransformer} & \underline{0.3937} & -1.1260 & -1.0573 & -1.1614 & \underline{93.73} \\
  \midrule
  Ours & \textbf{0.0368} & \textbf{-1.5305} & \textbf{-1.5718} & \textbf{-1.5642} & \textbf{97.52}\\ 
 \bottomrule
 \end{tabular}
 }
 \end{center}
\end{table}

\section{Related Works}

\subsection{Point Cloud Video Understanding}

Point cloud video contains complex spatial-temporal information and combines an intricate structure with both unordered (intra-frame) and ordered (inter-frame) nature. Early methods either simplify its structure by dimensionality reduction using projections \citep{luo2018fast}, or adopt voxelization to construct a regulated grid-based data \citep{Minkowski, wang20203dv}. Similar to projections/voxel-based methods in static point clouds, those methods also faced information loss and issues with processing efficiency. Recent methods \citep{liu2019meteornet, Min2020pointlstm, fan2021pstnet, fan21p4transformer} inclined to process the point cloud video directly with set abstraction \citep{qi2017pointnet++}. For instance, \cite{fan2021pstnet} proposed a 4D convolution that implicitly captures the dynamic of adjacent point cloud frames by adopting the set abstraction between them and processes the point cloud sequence recursively. After that, an improved version \citep{fan2022pstnet2} proposed to enhance dynamic capture with an additional temporal convolution. These point-based methods focus more on the motion representation and try to improve the dynamic capture process in different aspects. P4Transformer \citep{fan21p4transformer} and PST-Transformer \citep{fan2023psttransformer} captured dynamic by searching related points in the spatial-temporal space with attention-based networks. Kinet \citep{Zhong2022kinet} proposed a kinematics-inspired neural network and solved the dynamic capture in point cloud sequence using scene flow. Our Motion PointNet is built upon a brand-new perspective that treats the process of representation learning in point cloud video as a solvable PDE problem.

\subsection{PDE-Solving with Deep Models}

Our work is also related to solving PDE numerically with deep models. The PDE-solving problem has been widely explored with spectral methods since the last century \citep{gottlieb1977numerical, fornberg1998practical}. Recently, some research work explored the deep models for PDE due to their great nonlinear modeling capability \citep{li2020fourier, tran2021factorized,fanaskov2022spectral, liu2023htnet}. Additionally, PDE has garnered interest in vision research, finding applications in tasks like point cloud compression \citep{yang2023pde} and video prediction \citep{wu2023disentangling}. In this study, we endeavor to implement PDE in point cloud video understanding tasks. To this end, we have developed a novel Motion PointNet model utilizing PDE for the spatial-temporal correlations in point cloud video, thereby enhancing motion representation and raising performance in these tasks. To the best of our knowledge, this is the first application of PDE in this domain.

\section{Proposed Method}

\begin{figure}[tbp]
\begin{center}
\includegraphics[width=\linewidth]{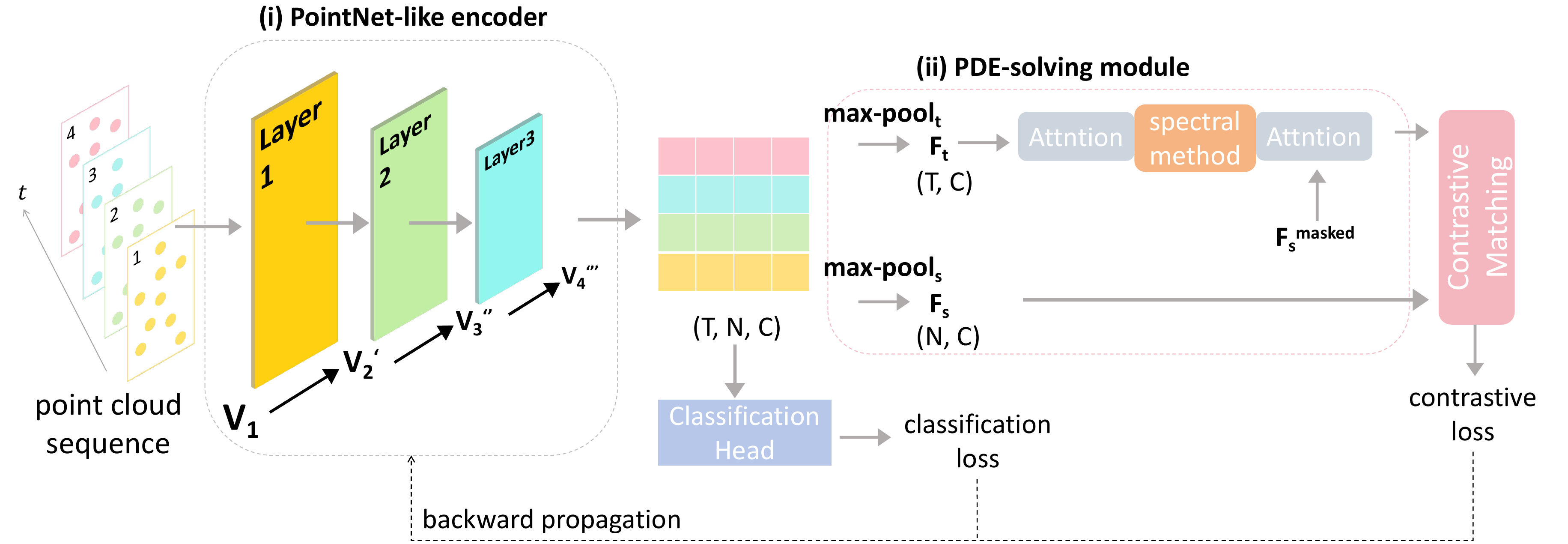}
\end{center}
\caption{Overall architecture of our Motion PointNet. \textbf{PointNet-like Encoder:} Benefiting from the rolling operation, as the network goes deeper, features from the current frame are continually aggregated to the next frame, while also perceiving more spatial information with a larger spatial receptive field. \textbf{PDE-solving module:} We then further refine the motion information by formulating this process as solvable PDE. The PDE-solving module provides additional supervision of the backbone with a cross-dimension feature reconstruction target.}
\label{fig:motionnet}
\end{figure}

We illustrate the proposed Motion PointNet for point cloud video in detail in the following sections. \Cref{fig:motionnet} shows the overall architecture of the Motion PointNet, which is composed of a PointNet-like encoder and a PDE-solving module. 

\subsection{PointNet-like Encoder}
\label{sec:encoder}

We design a lightweight yet effective encoder for the initial spatial variations in our Motion PointNet. Given a point cloud with $N$ points that presented as $P = \{p_1, p_2, ..., p_N\}$, where each point $p_i \in \mathbb{R}^3$ is specified by the geometric coordinates $\{x, y, z\}$. A point cloud video contains $T$ frames of point clouds presented as $V = \{P_1, P_2, ..., P_T\}$, combining characteristics of both unordered intra-frame and ordered inter-frame. Previous methods \citep{fan2021pstnet, fan21p4transformer, fan2023psttransformer} process the point cloud video $V$  recursively by sophisticated designed 4D convolutions. Differently, we process $T$ frames of data as batches following previous video networks \citep{lin2019tsm, wang2019event}, thus processing the point cloud video with the shape of $\{B \times T, N, C\}$. Here a `batch' represents a group of data entered into the network for training, and the `batch size' represents the number of training samples in each batch. Here the $batch~size = B \times T$. Usually, $C=3$ represents the spatial coordinates $\{x, y, z\}$. 

We then extend the spatial encoder from PointNet++ \citep{qi2017pointnet++} for static point clouds only to the temporal domain. For a static point cloud $P$, \cite{qi2017pointnet++} adopt a multilayer perceptron (MLP) for spatial set abstraction:

\begin{equation}
\label{eq:pointnet}
    Feature = \mathit{f}(P, P^{'}),
\end{equation}

where $\mathit{f}$ represents a standard PointNet++ layer. We omit some basic operations in point cloud processing to simplify the description (\textit{e.g.} sampling and grouping for the generation of $P^{'}$ from $P$). Here $P$ functions as the support points and $P^{'}$ is the downsampled query points from the original $P$. We recommend referring to PointNet++ \citep{qi2017pointnet++} for more details. \Cref{eq:pointnet} can be extended and further formed as follows when using batch processing, thus easily fitting our reshaped point cloud video:

\begin{equation}
\label{eq:batchpointnet}
    Feature = \mathit{f}(\{P_1, P_2, ...,\}, \{P_1, P_2, ...,\}^{'}) = \mathit{f}(V, V^{'}).
\end{equation}

However, the simple spatial encoder still lacks temporal consciousness and cannot well present the motion information. We solve this problem by adding a \textit{rolling} operation \textit{i.e.} \texttt{torch.roll()} on the temporal dimension which leads to frame misalignment in point cloud videos. In other words, we generate the support points and the query points from different point cloud frames instead of the same one to aggregate `motion' from $P_t \rightarrow P_{t+1}$:
\begin{equation}
\label{eq:pointcloudvideo}
    Feature = \mathit{f}(P_t, P_{t+1}^{'}).
\end{equation}

Here, the $P_t$ functions as the support points and $P_{t+1}^{'}$ is the downsampled query points from the next point cloud frame $P_{t+1}$. When using batch processing, \Cref{eq:pointcloudvideo} can be reformulated as follows:
\begin{equation}
\label{eq:batchpointcloudvideo}
    Feature = \mathit{f}(\{P_1, P_2, ..., P_t, ...\}, \{P_2, P_3, ..., P_{t+1}, ...\}^{'}) = \mathit{f}(V_1, V_{2}^{'}),
\end{equation}

where the index of $V$ represents the temporal index of the first point cloud frame in the point cloud sequence. In this way, the spatial set abstraction is extended to the temporal domain by operating on the adjacent frames while keeping its lightness and simplicity. We naturally stack multiple PointNet++ layers to build our PointNet-like encoder. As the network delves deeper, simultaneous temporal rolling and spatial abstraction persist, resulting in the expansion of the encoder's receptive fields in both spatial and temporal dimensions. Taking a 3-layer depth encoder as an example: 
\begin{equation}
\label{eq:encoder}
    Layer_1 = \mathit{f}(V_1, V_{2}^{'})\quad
    Layer_2 = \mathit{f}(V_{2}^{'}, V_{3}^{''})\quad
    Layer_3 = \mathit{f}(V_{3}^{''}, V_{4}^{'''}),
\end{equation}

where more superscript $'$ represents the larger spatial sampling scale than the previous layer. Different from previous methods, our encoder maintains the sequence length $T$ while aggregating temporal information from the current frame to the next frame, greatly enhancing the local information density of our features.

\subsection{PDE-solving Module} 

\paragraph{Building temporal-to-spatial mapping.}

\begin{figure}[htbp]
\begin{center}
\includegraphics[width=\linewidth]{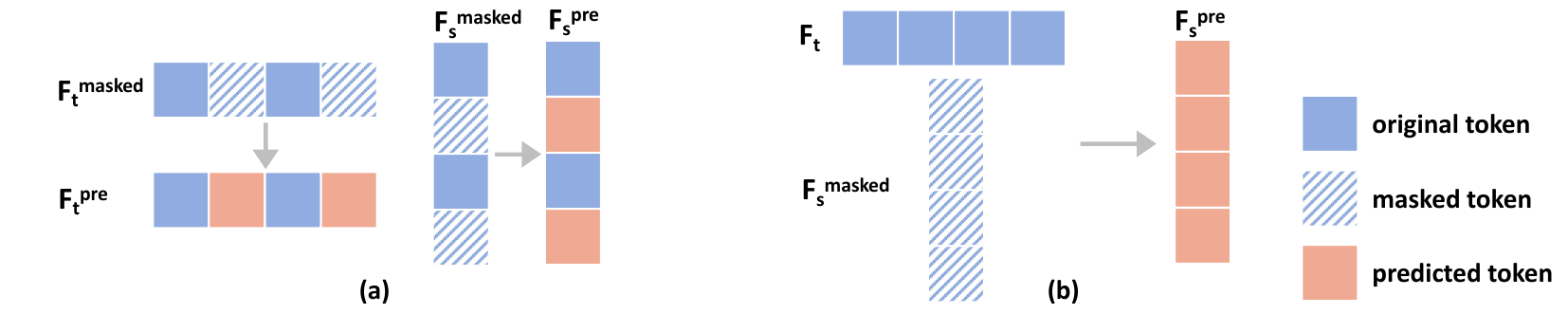}
\end{center}
\caption{Comparison between reconstruction target for (a) inner data distribution and (b) spatial-temporal correlation modeling. 
}
\label{fig:ads}
\end{figure}

Given the spatial-temporal feature acquired from the encoder $Feature \in \mathbb{R}^{T\times M}$, where $M < N$ is the number of spatial regions after aggregation. Subsequently, we proceed with max-pooling in both the temporal and spatial dimensions, respectively, and obtain the sub-global temporal feature $F_t \in \mathbb{R}^{T}$ and the sub-global spatial feature $F_s \in \mathbb{R}^{M}$. We target reconstructing the $F_s$ from a learnable parameters set $F_s^{masked}$ by reversing the $F_t$ with spectral methods (see \Cref{fig:ads} (b)). Both the $F_t$ and the $F_s$ are in the Banach spaces $\mathcal{F} = \mathcal{F}(\mathcal{D};\mathbb{R}^{d_F})$, where $\mathcal{D}\subset \mathbb{R}^{d}$ is a bounded open set. Based on assumptions from \cite{lu2021learning} and \cite{li2021fourier}, we can solve the PDE with a deep model $\mathcal{M}_{\theta}$ by approximating the optimal operator. This process can be formulated as follows:
\begin{equation}
\label{eq:pdesolve}
    \mathcal{M}: F_t \stackrel{\theta}{\longrightarrow} F_s,
\end{equation}

where $\theta$ is the learnable parameter set. Our PDE-solving module directly incorporates the feature variations over time as the variable $\theta$ in the modeling process. This allows the PDE to capture a more accurate and nuanced representation of how point cloud features evolve compared to previous temporal modeling approaches. Our approach also differs from the existing reconstruction-based method that mainly addresses the inner data distribution patterns by reconstructing masked tokens (see \Cref{fig:ads} (a)). These methods focus on either the spatial or temporal dimension solely and are usually sensitive to mask ratio. Furthermore, simply recovering the masked data does not meet our purposes of spatial-temporal correlation modeling. 

\paragraph{Solving PDE mapping.}

We then introduce the core of the PDE-solving module with a spectral method. The approximating of $\theta$ in \Cref{eq:pdesolve} can be formulated as follows:

\begin{equation}
\label{eq:operators}
    \mathcal{M}_{\theta} = \sum_{i=1}^{O} \mathit{w}_i \mathcal{M}_{\theta, i},
\end{equation}

where $O$ is the number of operators and $\mathit{w}_i$ is learnable weight. As shown in \Cref{fig:motionnet} (iii), the core of the PDE-solving module is composed of a multi-head self-attention (MHSA) layer, a spectral method layer, and a multi-head cross-attention (MHCA) layer. Although a simplistic deep model \citep{liu2023htnet} can be used for PDE-solving, the design attempts to learn the operator as a whole ($O=1$) while still maintaining network efficiency is challenging due to the complexities of input-output mappings in high-dimensional space \citep{wu2023LSM, karniadakis2021physics}. We tackle this problem with the combination of the attention mechanism and the classic spectral method \citep{tolstov2012fourier} for PDE, which decomposes complex nonlinear mappings into multiple basis operators, while also holding the universal approximation capacity with theoretical guarantees.

For every $\mathbf{f_t} \in F_t$, we use trigonometric as the basis operators following \cite{lu2021learning} and \cite{li2021fourier}:

\begin{gather}
\label{eq:bias}
    \mathcal{M}_{\theta, (2k-1)}(\mathbf{f_t}) = sin(k\mathbf{f_t}), \ \mathcal{M}_{\theta, (2k)}(\mathbf{f_t}) = cos(k\mathbf{f_t}), \
    k \in \{1, ..., N/2\}
\end{gather}

where $N$ is even. Then, the calculation of the mapping output can be formulated as follows:

\begin{gather}
\label{eq:output}
    F_{t\rightarrow s} = F_t + \mathit{w}_{sin}[\mathcal{M}_{\theta, (2k-1)}(\mathbf{F_t})]_{k=1}^{O/2} + \mathit{w}_{cos}[\mathcal{M}_{\theta, (2k)}(\mathbf{F_t})]_{k=1}^{O/2}.
\end{gather} 

Building upon the spectral method we elaborate above, we use a standard MHSA \citep{vaswani2017attention} layer to enhance the temporal feature before feeding into the spectral method layer. Furthermore, we align the two different Banach spaces of the mapping output $F_{t\rightarrow s}$ and $F_s^{masked}$ by an MHCA layer and output the predicted $\hat{F_s}$. \footnote{We detail this process in \Cref{detailsMHSA}} The $F_s^{masked}$ is initialized with the same shape as $F_s$. 

\paragraph{Contrastive Matching Loss.}

We compare the predicted $\hat{F_s}$ with the ground truth $F_s$ using a contrastive-based InfoNCE loss \citep{oord2018representation} to refine the modeling of the spatial-temporal correlations to a distinct objective. We consider both the input $F_t$ and the output $F_s$ to contain more abstract information about the spatio-temporal features. By treating the $\mathbf{f}_{-}$ in $Feature \in \mathbb{R}^{T\times M}$ before applying spatial/temporal pooling as the negative sample, we force the model to learn the implicit mapping between the spatial and temporal space, instead of reconstructing the de-pooling feature by closer both Ft and Fs to $Feature$ (which is against the learning objectives). For every $\hat{\mathbf{f_s}} \in \hat{F_s}$, we treat the corresponding token in $F_s$ as positive sample $\mathbf{f_s}_{+}$. The loss function can be formulated as follows:

\begin{equation}
\label{eq:loss}
    \mathcal{L} = \sum_{\hat{\mathbf{f_s}}_{,i}\in \hat{F_s}}-log\frac{exp(\hat{\mathbf{f_s}}_{,i}^T\mathbf{f_s}_{+}/\tau)}{exp(\hat{\mathbf{f_s}}_{,i}^T\mathbf{f_s}_{+}/\tau) + \sum_{\mathbf{f}_{-,j}\in Feature}exp(\hat{\mathbf{f_s}}_{,i}^T\mathbf{f}_{-,j}/\tau)},  
\end{equation}

where $\tau$ is a temperature that controls the network sensitivity to positive and negative samples. Several ablation studies in Section \Cref{sec:Ablation} underscore the superiority of our loss design.

\section{Experiment}

\subsection{Experimental Settings}

We evaluate the proposed Motion PointNet on three benchmarks including the MSRAction-3D \citep{li2010msraction} dataset, NTU RGB+D \citep{shahroudy2016ntu} dataset and UTD-MHAD \citep{utddataset} dataset for point cloud video action recognition. We conduct all our experiments on the NVIDIA A100 GPUs. Following the previous works \citep{Zhong2022kinet,fan2021pstnet,fan2023psttransformer, utddataset}, we use default data splits in all evaluated datasets for fair comparisons.

\paragraph{Training Settings.} We evaluate our Motion PointNet using several training strategies including end-to-end training, pre-train and fine-tuning, and linear probing. Leveraging the contrastive learning mechanism, we found that pre-train and fine-tuning can bring the most improvement in \Cref{sec:Ablation} and report them as our main results in the following sections. We pre-train our Motion PointNet with the PDE-solving module and only fine-tune the encoder we proposed in \Cref{sec:encoder} with a classification head. We report results from the fine-tuned models. We further investigate the self-supervised potential of the PDE-solving module on other backbone models.

\begin{table*}[htbp]
\caption{Comparison with current state-of-the-art on MSRAction-3D dataset.}
\label{tab:msr}
\begin{center}
\begin{tabular}{ccccc>{\columncolor{lightgray!40}}c}
  \toprule
    Methods & \multicolumn{5}{c}{Accuracy(\%) of different frame rate} \\
    \midrule
    & 4-frame & 8-frame & 12-frame & 16-frame & 24-frame \\
    MeteorNet \citep{liu2019meteornet} & 78.11 & 81.14 & 86.53 & 88.21 & 88.50 \\
    P4Transformer \citep{fan21p4transformer} & 80.13 & 83.17 & 87.54 & 89.56 & 90.94 \\
    PSTNet \citep{fan2021pstnet} & 81.14 & 83.50 & 87.88 & 89.90 & 91.20 \\
    SequentialPointNet \citep{li2021sequentialpointnet} & 77.66 & 86.45 & 88.64 & 89.56 & 91.94 \\
    PointMapNet \citep{pointmapnet} & 79.04 & 84.93 &  87.13 & 89.71 & 91.91 \\
    PSTNet++ \citep{fan2022pstnet2} & 81.53 & 83.50 & 88.15 & 90.24 & 92.68 \\
    Kinet \citep{Zhong2022kinet} & 79.80 & 83.84 & 88.53 & 91.92 & 93.27 \\
    3DInAction \citep{ben20233dinaction} & 80.47 & 86.20 & 88.22 & 90.57 & 92.23 \\
    PST-Transformer \citep{fan2023psttransformer} & 81.14 & 83.97 & 88.15 & 91.98 & 93.73 \\
  \midrule
    \textbf{Motion PointNet} & 79.46 & 85.88 & 90.57  & 93.33 & \textbf{97.52} \\
 \bottomrule
 \end{tabular}
 \end{center}
\end{table*}

\subsection{Action Recognition Results}

\textbf{MSRAction-3D} dataset includes 567 depth map sequences of 20 action classes performed by 10 subjects. To generate point cloud videos from the original data, we adopt the standard method following \cite{liu2019meteornet} and \cite{fan2021pstnet,fan21p4transformer}, and report the average accuracy of our experiment over 10 runs following the convention. We compare our Motion PointNet to prior works in \Cref{tab:msr}. Our report proves that the proposed method outperforms the current SOTA by significant margins, gaining a \textbf{+3.79\%} accuracy with 24-frame input. Furthermore, it maintains superior performance with reduced frame input (12/16-frame), demonstrating the robustness of the proposed Motion PointNet. The performance on 4/8-frame MSRAction-3D indicates a slight limitation on short video input with a comparable accuracy. Indicate that the simple and explicit temporal informativeness of adjacent frames is proportional to the length of the input video.

\begin{table*}[htbp]
 \caption{Qualitative results for efficiency evaluation on MSRAction-3D. Notice that the reported runtime results are on 24-frame MSRAction-3D.}
\label{tab:parm}
\begin{center}
{
\begin{tabular}{lcccccc}
  \toprule
    Methods & flames & FLOPs(G) & Param.(M) & Acc.(\%) & time(ms) \\
    \midrule
    PSTNet \citep{fan2021pstnet} & \multirow{5}{*}{16} & 54.09 & 8.44 & 89.90 & 63.88 \\
    MeteorNet \citep{liu2019meteornet} & & 1.70 & 17.60 & 88.21 & 80.11 \\
    P4Transformer \citep{fan21p4transformer} & & 40.38 & 42.07 & 89.56 & 25.18\\
    PST-Transformer \citep{fan2023psttransformer} & & - & 44.20 & 91.98 & 69.37\\
    Kinet \citep{Zhong2022kinet} & & 10.35 & 3.20 & 91.92 & - \\
    \midrule
    \textbf{Motion PointNet} & 16/24 & \textbf{0.55}/0.82 & \textbf{0.72} & 93.33/\textbf{97.52} & \textbf{1.17} \\
 \bottomrule
 \end{tabular}
 }
 \end{center}
\end{table*}

Notably, the proposed method not only attains state-of-the-art performance but also surpasses existing models in terms of model parameters, complexity, and running time. As illustrated in \Cref{tab:parm}, previous approaches that rely on sophisticated 4D convolutions are usually accompanied by intricate computational demands and substantial learning parameters. For instance, PSTNet \citep{fan2021pstnet} is composed of a hierarchical architecture with 4-layer 4D convolutions, which result in over 50G FLOPs. When further improving the performance by even more complex networks \citep{fan21p4transformer, fan2023psttransformer}, the learning parameters reach a staggering 40M+. Differently, our Motion PointNet miniaturizes the model in both FLOPs and learning parameters to the SOTA level. Remarkably, our Motion PointNet surpasses the current state-of-the-art with a \textbf{0.55}G FLOPs (when comparing at a 16-frame input) and \textbf{0.72}M learning parameters. We further visualize the points corresponding to high feature response in \Cref{apx:vis}, please check for details. 

\begin{wraptable}{r}{0.55\linewidth}
\vspace{-.6cm}
\caption{Compare to the current state-of-the-art on NTU RGB+D dataset.}
\label{tab:ntu60}
\begin{center}
\resizebox{\linewidth}{!}
{
\begin{tabular}{lccc}
  \toprule
  \multirow{2}{*}{Methods} & \multirow{2}{*}{Modalities} & Cross  &  Cross \\
  & & Subject & View \\
    \midrule
    Li et al. \citep{li2018unsupervised} & \multirow{3}{*}{depth map} & 68.1 & 83.4 \\
    Wang et	al. \citep{wang2018depth} & & 87.1 & 84.2 \\
    MVDI \citep{xiao2019mvdi} & & 84.6 & 87.3 \\
    \midrule
    SkeleMotion \citep{caetano2019skelemotion} & \multirow{3}{*}{skeleton} & 69.6 & 80.1 \\
    DGNN \citep{shi2019dgnn} & & 89.9 & 96.1 \\
    MS-G3D \citep{liu2020msg3d} & & 91.5 & 96.2 \\
    \midrule
    3DV \citep{wang20203dv} & \multirow{8}{*}{points} & 88.8 & 96.3 \\
    P4Transformer \citep{fan21p4transformer} & & 90.2 & 96.4 \\
    PST-Transformer \citep{fan2023psttransformer} & & 91.0 & 96.4 \\
    Kinet \citep{Zhong2022kinet} & & 92.3 & 96.4 \\
    PSTNet \citep{fan2021pstnet} & & 90.5 & 96.5 \\
    PSTNet++ \citep{fan2022pstnet2} & & 91.4 & 96.7 \\
    PointMapNet \citep{pointmapnet} & & 89.4 & 96.7 \\
    SequentialPointNet \citep{li2021sequentialpointnet} & & 90.3 & 97.6 \\
  \midrule
    \textbf{Motion PointNet} & points & \textbf{92.9} & \textbf{98.0} \\
 \bottomrule
 \end{tabular}
 }
 \end{center}
\end{wraptable}

\textbf{NTU RGB+D} dataset contains 60 action classes and 56,880 video samples, which is a large-scale dataset consisting of complex scenes with noisy background points. We report the results of the cross-subject and cross-view scenarios following the official data partition \citep{shahroudy2016ntu}. We compare our Motion PointNet to prior works in \Cref{tab:ntu60}. Our report proves that the proposed method maintains its superiority on the large-scale dataset. Our Motion PointNet is superior to most of the methods with different input modalities including depth map, skeleton, and dense points. It consistently outperforms the large model including PSTNet \citep{fan2021pstnet}, P4Transformer \citep{fan21p4transformer}, and PST-Transformer \citep{fan2023psttransformer} in both the cross-subject (92.9\% accuracy) and cross-view (98.0\% accuracy) protocols, while forming a way more lightweight network. As shown in \Cref{tab:ntuparm}, our Motion PointNet offers consistent superiority in lightweight regarding model parameters (1.64M parameters) and computational complexity (15.47G FLOPs).

\begin{table*}[htbp]
\caption{Qualitative results for efficiency evaluation on NTU RGB+D dataset.}
\label{tab:ntuparm}
\begin{center}
{
  \begin{tabular}{lcc}
    \toprule
    Model & FLOPs(G) & PARAMS(M) \\
    \midrule
    PSTNet \citep{fan2021pstnet} & 19.58 & 8.52 \\
    PointMapNet \citep{pointmapnet} & - & 2.65 \\
    P4Transformer \citep{fan21p4transformer} & 48.63 & 65.17 \\
    PST-Transformer \citep{fan2023psttransformer} & 48.68 & 65.19 \\
    GeometryMotion-Net \citep{liu2021GeometryMotionnet} & 68.42 & 40.44 \\
    \midrule
    \textbf{Motion PointNet} & \textbf{15.47} & \textbf{1.64} \\
    \bottomrule
  \end{tabular}
}
\end{center}
\vspace{.5cm}
\end{table*}

\begin{table}[htbp]
\caption{Hyper-settings of our Motion PointNet.}
\label{encodersetting}
\centering
{\begin{tabular}{lccc}
    \toprule
    \multicolumn{2}{c}{Dataset} & MSRAction-3D &  NTU RGB+D \\
    \midrule
    & Input points $\times$ frames  & 2048 $\times$ 24 & 2048 $\times$ 24\\
    \multirow{2}{*}{Encoder} & Number of encoder layers & 3 & 5 \\
    \multirow{2}{*}{Setting} & Spatial stride & 32, 8, 2 & 8, 8, 1, 1, 4 \\
    & K-neighbors & 48, 32, 8 & 32, 48, 16, 24, 32 \\
    & Output feature channel & 1024 & 1024 \\
    \midrule
    \multirow{2}{*}{PARAMS(M)} & w/o PDE-solving & 0.72 & 1.64 \\
    & w/ PDE-solving & 5.95 & 6.83 \\  
    \midrule
    \multirow{2}{*}{FLOPs(G)} & w/o PDE-solving & 0.82 & 15.47 \\
    & w/ PDE-solving & 1.06 & 15.73 \\
    \bottomrule
  \end{tabular}
}
\end{table}

We also report the hyper-settings of our Motion PointNet for the two aforementioned datasets in \Cref{encodersetting}. We modified the basis hyperparameters in the table and selected the best-setting group in our experiments. The NTU RGB+D dataset requires a deeper network due to its complex scenes and large scale. Notice that our PDE-solving module is also lightweight with only additional +5.2M parameters and +0.2G FLOPs in both settings.

\begin{wraptable}{l}{0.5\linewidth}
\vspace{-.5cm}
\setlength{\belowcaptionskip}{.3cm}
\caption{Compare to the current state-of-the-art on UTD-MHAD benchmark.}
\label{tab:utd}
\centering
{
\begin{tabular}{lc}
\toprule
Methods & Accuracy(\%)  \\
\midrule
SequentialPointNet \citep{li2021sequentialpointnet} & 92.31 \\
PointMapNet \citep{pointmapnet} & 91.61 \\
\midrule
Motion PointNet & 92.79 \\
\bottomrule   
\end{tabular}
}
\vspace{-.3cm}
\end{wraptable}

\textbf{UTD-MHAD} dataset contains 27 classes and 861 data sequences for action recognition. We apply our Motion PointNet to the UTD-MHAD benchmark and compare the proposed approach with current SOTA methods. The encoder settings of the Motion PointNet are consistent with the settings for the NTU RGB+D benchmark. Results in \Cref{tab:utd} illustrate the accuracy of different approaches. Our Motion PointNet maintains its superior performance with the highest accuracy of 92.79\%.

\subsection{Ablation studies}
\label{sec:Ablation}

\begin{wraptable}{r}{0.55\linewidth}
\vspace{-.6cm}
\setlength{\belowcaptionskip}{.3cm}
\caption{Ablation of different training settings for the Motion PointNet. Experiments are conducted on the MSRAction-3D benchmark with 24 frames.}
\label{tab:settings}
\begin{center}
{\begin{tabular}{lr}
\toprule
Settings & Accuracy(\%)  \\
\midrule
pretrain + finetune & 97.52 \\
End-to-end & $-$ 0.64 \\
Linear probing & $-$ 1.41 \\
\bottomrule
\end{tabular}}
\end{center}
\vspace{-.cm}
\end{wraptable}

\paragraph{Ablation on Training Settings.}

Our pre-train and fine-tuning training process is designed to first allow the network to learn a robust spatial-temporal representation under the guidance of our PDE-solving module, and then to refine this representation using the classification head in the fine-tuning stage. In this way, we can keep our encoder as lightweight as possible while still enforcing its strong learning ability of the spatial-temporal correlations of the point cloud video data. To further demonstrate the effectiveness of our proposed method, we conduct two baseline comparisons in \Cref{tab:settings}. The experiments are conducted on the MSR-Action3D Dataset with 24 frames.

Here end-to-end training indicates training our Motion PointNet (encoder + PDE-solving module) with the classification head using the same number of iterations of the pre-train and fine-tuning training. Linear probing indicates fine-tuning the classification head while freezing the PointNet-like encoder to evaluate the pre-trained representation.

\begin{wraptable}{r}{0.55\linewidth}
\vspace{-.6cm}
\setlength{\belowcaptionskip}{.3cm}
\caption{Ablation on PDE-solving module. Experiments are conducted on the MSRAction-3D benchmark.}
\label{tab:ads}
\centering
{\begin{tabular}{lc}
\toprule
  Methods & Accuracy(\%)  \\
    \midrule
    PST-Transformer \citep{fan2023psttransformer} & 93.73 \\
     + PDE-solving          & 95.05 (\textbf{+1.32}) \\
  \midrule
    Our Encoder  & 95.76 \\
     + PDE-solving          & 97.52 (\textbf{+1.76}) \\
 \bottomrule
\end{tabular}}
\vspace{-.3cm}
\end{wraptable}

We also conduct extensive ablation experiments on the proposed Motion PointNet. Results in \Cref{tab:ads} show that the proposed encoder itself has outperformed the PST-Transformer \citep{fan2023psttransformer} with a 95.76\% accuracy. Furthermore, the PDE-solving module brings a significant improvement (+1.75\% accuracy) to our encoder. We also implement our PDE-solving module on the PST-Transformer. After pertaining together with our PDE-solving module, the PST-Transformer encoder also achieved a +1.32\% accuracy improvement. This underscores the universality of our PDE-solving module and the applicability of the PDE-solving perspective across different scenarios in point cloud video action recognition.

We further validate the effectiveness of different components in the PDE-solving module. Results are shown in \Cref{tab:components}. Firstly, we assess the individual contributions of the three layers that constitute the core of PDE-solving module. We observe that the spectral method primarily contributes to the performance enhancement, with the MHSA and MHCA layers also demonstrating their indispensability. The following results prove the utility of our contrastive matching loss. Other measures including cosine similarity and L2 distance are not ideal when we want to maximize the similarity between representations since they are either insensitive to linear scale or unbounded and harder to optimize. Finally, we validate different reconstruction targets in the PDE-solving module. The performance decays when we attempt to model inversely by solving $F_s \rightarrow F_t$ instead of $F_t \rightarrow F_s$. We hypothesize that this phenomenon arises because $F_t$ preserves a higher degree of integrated temporal information and the PDE is more explanatory with the $F_t \rightarrow F_s$ target.

\begin{table}[htbp]
\caption{Ablation of different components of PDE-solving module. Experiments are conducted on the MSRAction-3D benchmark.}
\label{tab:components}
\begin{center}
{\begin{tabular}{lcr}
\toprule
\multicolumn{2}{c}{Settings} & Accuracy(\%)  \\
\midrule
\multicolumn{2}{c}{full PDE-solving module} & 97.52 \\
\midrule
\multirow{3}{*}{PDE-solving core} & w/o MHSA & $-$ 0.71 \\
& w/o spectral method & $-$ 1.14 \\
& w/o MHCA & $-$ 0.41 \\
\midrule
\multirow{2}{*}{w/o contrastive matching} &  L2 similarity & $-$ 0.63 \\
&  Cosine similarity & $-$ 0.43 \\
w/ contrastive matching &  InfoNCE loss & $\pm$ 0. \\
\midrule
\multirow{2}{*}{reconstruction targets} & $F_s \rightarrow F_t$ & $-$ 0.68 \\
& $F_t \rightarrow F_s$ & $\pm$ 0. \\
\bottomrule
\end{tabular}}
\end{center}
\end{table}

\section{Conclusion}

We have presented a novel architecture called Motion PointNet for point cloud video representation learning through PDE-solving. Our approach leverages a lightweight PointNet-like encoder and a specialized PDE-solving module to establish a unified spatio-temporal representation, enhancing point cloud video learning significantly. Our extensive experiments across multiple benchmarks demonstrate the superiority of Motion PointNet, achieving state-of-the-art accuracy with notably reduced model complexity. This work not only advances the field of point cloud video representation learning but also opens new avenues for future research on efficient and effective motion capture in point clouds. The remarkable results underscore the potential of PDE-based solutions in addressing complex spatio-temporal data challenges. In future work, we aim to further extend our Motion PointNet to an extensive range of point cloud video understanding tasks including segmentation, detection, and object tracking.

\bibliography{arxiv_main}

\begin{thebibliography}{}

\bibitem[Ben-Shabat et~al., 2023]{ben20233dinaction}
Ben-Shabat, Y., Shrout, O., and Gould, S. (2023).
\newblock 3dinaction: Understanding human actions in 3d point clouds.
\newblock {\em arXiv preprint arXiv:2303.06346}.

\bibitem[Caetano et~al., 2019]{caetano2019skelemotion}
Caetano, C., Sena, J., Br{\'e}mond, F., Dos~Santos, J.~A., and Schwartz, W.~R. (2019).
\newblock Skelemotion: A new representation of skeleton joint sequences based on motion information for 3d action recognition.
\newblock In {\em 2019 16th IEEE international conference on advanced video and signal based surveillance (AVSS)}, pages 1--8. IEEE.

\bibitem[Chen et~al., 2015]{utddataset}
Chen, C., Jafari, R., and Kehtarnavaz, N. (2015).
\newblock Utd-mhad: A multimodal dataset for human action recognition utilizing a depth camera and a wearable inertial sensor.
\newblock In {\em 2015 IEEE International Conference on Image Processing (ICIP)}, pages 168--172.

\bibitem[Choy et~al., 2019]{Minkowski}
Choy, C., Gwak, J., and Savarese, S. (2019).
\newblock 4d spatio-temporal convnets: Minkowski convolutional neural networks.
\newblock In {\em Proceedings of the IEEE/CVF Conference on Computer Vision and Pattern Recognition}, pages 3075--3084.

\bibitem[Fan et~al., 2021a]{fan21p4transformer}
Fan, H., Yang, Y., and Kankanhalli, M. (2021a).
\newblock Point 4d transformer networks for spatio-temporal modeling in point cloud videos.
\newblock In {\em {IEEE/CVF} Conference on Computer Vision and Pattern Recognition, {CVPR}}.

\bibitem[Fan et~al., 2023]{fan2023psttransformer}
Fan, H., Yang, Y., and Kankanhalli, M. (2023).
\newblock Point spatio-temporal transformer networks for point cloud video modeling.
\newblock {\em IEEE Transactions on Pattern Analysis and Machine Intelligence}, 45(2):2181--2192.

\bibitem[Fan et~al., 2021b]{fan2021pstnet}
Fan, H., Yu, X., Ding, Y., Yang, Y., and Kankanhalli, M. (2021b).
\newblock Pstnet: Point spatio-temporal convolution on point cloud sequences.
\newblock In {\em International Conference on Learning Representations}.

\bibitem[Fan et~al., 2022]{fan2022pstnet2}
Fan, H., Yu, X., Yang, Y., and Kankanhalli, M. (2022).
\newblock Deep hierarchical representation of point cloud videos via spatio-temporal decomposition.
\newblock {\em IEEE Transactions on Pattern Analysis and Machine Intelligence}, 44(12):9918--9930.

\bibitem[Fanaskov and Oseledets, 2022]{fanaskov2022spectral}
Fanaskov, V. and Oseledets, I. (2022).
\newblock Spectral neural operators.
\newblock {\em arXiv preprint arXiv:2205.10573}.

\bibitem[Fornberg, 1998]{fornberg1998practical}
Fornberg, B. (1998).
\newblock {\em A practical guide to pseudospectral methods}.
\newblock Number~1. Cambridge university press.

\bibitem[Gottlieb and Orszag, 1977]{gottlieb1977numerical}
Gottlieb, D. and Orszag, S.~A. (1977).
\newblock {\em Numerical analysis of spectral methods: theory and applications}.
\newblock SIAM.

\bibitem[Karniadakis et~al., 2021]{karniadakis2021physics}
Karniadakis, G.~E., Kevrekidis, I.~G., Lu, L., Perdikaris, P., Wang, S., and Yang, L. (2021).
\newblock Physics-informed machine learning.
\newblock {\em Nature Reviews Physics}, 3(6):422--440.

\bibitem[Khosla et~al., 2020]{khosla2020supervised}
Khosla, P., Teterwak, P., Wang, C., Sarna, A., Tian, Y., Isola, P., Maschinot, A., Liu, C., and Krishnan, D. (2020).
\newblock Supervised contrastive learning.
\newblock {\em Advances in neural information processing systems}, 33:18661--18673.

\bibitem[Li et~al., 2018]{li2018unsupervised}
Li, J., Wong, Y., Zhao, Q., and Kankanhalli, M.~S. (2018).
\newblock Unsupervised learning of view-invariant action representations.
\newblock {\em Advances in neural information processing systems}, 31.

\bibitem[Li et~al., 2010]{li2010msraction}
Li, W., Zhang, Z., and Liu, Z. (2010).
\newblock Action recognition based on a bag of 3d points.
\newblock In {\em 2010 IEEE Computer Society Conference on Computer Vision and Pattern Recognition - Workshops}, pages 9--14.

\bibitem[Li et~al., 2021a]{li2021sequentialpointnet}
Li, X., Huang, Q., Wang, Z., Hou, Z., and Yang, T. (2021a).
\newblock Sequentialpointnet: A strong parallelized point cloud sequence network for 3d action recognition.
\newblock {\em arXiv preprint arXiv:2111.08492}.

\bibitem[Li et~al., 2023]{pointmapnet}
Li, X., Huang, Q., Zhang, Y., Yang, T., and Wang, Z. (2023).
\newblock Pointmapnet: Point cloud feature map network for 3d human action recognition.
\newblock {\em Symmetry}, 15(2).

\bibitem[Li et~al., 2020]{li2020fourier}
Li, Z., Kovachki, N., Azizzadenesheli, K., Liu, B., Bhattacharya, K., Stuart, A., and Anandkumar, A. (2020).
\newblock Fourier neural operator for parametric partial differential equations.
\newblock {\em arXiv preprint arXiv:2010.08895}.

\bibitem[Li et~al., 2021b]{li2021fourier}
Li, Z., Kovachki, N.~B., Azizzadenesheli, K., liu, B., Bhattacharya, K., Stuart, A., and Anandkumar, A. (2021b).
\newblock Fourier neural operator for parametric partial differential equations.
\newblock In {\em International Conference on Learning Representations}.

\bibitem[Lin et~al., 2019]{lin2019tsm}
Lin, J., Gan, C., and Han, S. (2019).
\newblock Tsm: Temporal shift module for efficient video understanding.
\newblock In {\em Proceedings of the IEEE/CVF international conference on computer vision}, pages 7083--7093.

\bibitem[Liu et~al., 2022]{liu2022GeometryMotionTransformer}
Liu, J., Guo, J., and Xu, D. (2022).
\newblock Geometrymotion-transformer: An end-to-end framework for 3d action recognition.
\newblock {\em IEEE Transactions on Multimedia}, pages 1--13.

\bibitem[Liu and Xu, 2021]{liu2021GeometryMotionnet}
Liu, J. and Xu, D. (2021).
\newblock Geometrymotion-net: A strong two-stream baseline for 3d action recognition.
\newblock {\em IEEE Transactions on Circuits and Systems for Video Technology}, 31(12):4711--4721.

\bibitem[Liu et~al., 2023]{liu2023htnet}
Liu, X., Xu, B., and Zhang, L. (2023).
\newblock {HT}-net: Hierarchical transformer based operator learning model for multiscale {PDE}s.

\bibitem[Liu et~al., 2019]{liu2019meteornet}
Liu, X., Yan, M., and Bohg, J. (2019).
\newblock Meteornet: Deep learning on dynamic 3d point cloud sequences.
\newblock In {\em ICCV}.

\bibitem[Liu et~al., 2020]{liu2020msg3d}
Liu, Z., Zhang, H., Chen, Z., Wang, Z., and Ouyang, W. (2020).
\newblock Disentangling and unifying graph convolutions for skeleton-based action recognition.
\newblock In {\em Proceedings of the IEEE/CVF conference on computer vision and pattern recognition}, pages 143--152.

\bibitem[Lu et~al., 2021]{lu2021learning}
Lu, L., Jin, P., Pang, G., Zhang, Z., and Karniadakis, G.~E. (2021).
\newblock Learning nonlinear operators via deeponet based on the universal approximation theorem of operators.
\newblock {\em Nature machine intelligence}, 3(3):218--229.

\bibitem[Luo et~al., 2018]{luo2018fast}
Luo, W., Yang, B., and Urtasun, R. (2018).
\newblock Fast and furious: Real time end-to-end 3d detection, tracking and motion forecasting with a single convolutional net.
\newblock In {\em Proceedings of the IEEE conference on Computer Vision and Pattern Recognition}, pages 3569--3577.

\bibitem[Min et~al., 2020]{Min2020pointlstm}
Min, Y., Zhang, Y., Chai, X., and Chen, X. (2020).
\newblock An efficient pointlstm for point clouds based gesture recognition.
\newblock In {\em Proceedings of the IEEE/CVF Conference on Computer Vision and Pattern Recognition (CVPR)}.

\bibitem[Oord et~al., 2018]{oord2018representation}
Oord, A. v.~d., Li, Y., and Vinyals, O. (2018).
\newblock Representation learning with contrastive predictive coding.
\newblock {\em arXiv preprint arXiv:1807.03748}.

\bibitem[Pu et~al., 2022]{shi2022Representation}
Pu, S., Zhao, K., and Zheng, M. (2022).
\newblock Alignment-uniformity aware representation learning for zero-shot video classification.
\newblock In {\em 2022 IEEE/CVF Conference on Computer Vision and Pattern Recognition (CVPR)}, pages 19936--19945.

\bibitem[Qi et~al., 2017]{qi2017pointnet++}
Qi, C.~R., Yi, L., Su, H., and Guibas, L.~J. (2017).
\newblock Pointnet++: Deep hierarchical feature learning on point sets in a metric space.
\newblock {\em Advances in neural information processing systems}, 30.

\bibitem[Shahroudy et~al., 2016]{shahroudy2016ntu}
Shahroudy, A., Liu, J., Ng, T.-T., and Wang, G. (2016).
\newblock Ntu rgb+ d: A large scale dataset for 3d human activity analysis.
\newblock In {\em Proceedings of the IEEE conference on computer vision and pattern recognition}, pages 1010--1019.

\bibitem[Shi et~al., 2019]{shi2019dgnn}
Shi, L., Zhang, Y., Cheng, J., and Lu, H. (2019).
\newblock Skeleton-based action recognition with directed graph neural networks.
\newblock In {\em 2019 IEEE/CVF Conference on Computer Vision and Pattern Recognition (CVPR)}, pages 7904--7913.

\bibitem[Song et~al., 2022]{song2022pref}
Song, L., Gong, X., Planche, B., Zheng, M., Doermann, D., Yuan, J., Chen, T., and Wu, Z. (2022).
\newblock Pref: Predictability regularized neural motion fields.
\newblock In {\em European Conference on Computer Vision}, pages 664--681. Springer.

\bibitem[Tolstov, 2012]{tolstov2012fourier}
Tolstov, G.~P. (2012).
\newblock {\em Fourier series}.
\newblock Courier Corporation.

\bibitem[Tran et~al., 2021]{tran2021factorized}
Tran, A., Mathews, A., Xie, L., and Ong, C.~S. (2021).
\newblock Factorized fourier neural operators.
\newblock {\em arXiv preprint arXiv:2111.13802}.

\bibitem[Vaswani et~al., 2017]{vaswani2017attention}
Vaswani, A., Shazeer, N., Parmar, N., Uszkoreit, J., Jones, L., Gomez, A.~N., Kaiser, {\L}., and Polosukhin, I. (2017).
\newblock Attention is all you need.
\newblock In {\em Advances in neural information processing systems}, pages 5998--6008.

\bibitem[Wang et~al., 2018]{wang2018depth}
Wang, P., Li, W., Gao, Z., Tang, C., and Ogunbona, P.~O. (2018).
\newblock Depth pooling based large-scale 3-d action recognition with convolutional neural networks.
\newblock {\em IEEE Transactions on Multimedia}, 20(5):1051--1061.

\bibitem[Wang et~al., 2019]{wang2019event}
Wang, Q., Zhang, Y., Yuan, J., and Lu, Y. (2019).
\newblock Space-time event clouds for gesture recognition: From rgb cameras to event cameras.
\newblock In {\em 2019 IEEE Winter Conference on Applications of Computer Vision (WACV)}, pages 1826--1835.

\bibitem[Wang and Isola, 2020]{wang2020understanding}
Wang, T. and Isola, P. (2020).
\newblock Understanding contrastive representation learning through alignment and uniformity on the hypersphere.
\newblock In {\em International Conference on Machine Learning}, pages 9929--9939. PMLR.

\bibitem[Wang et~al., 2020]{wang20203dv}
Wang, Y., Xiao, Y., Xiong, F., Jiang, W., Cao, Z., Zhou, J.~T., and Yuan, J. (2020).
\newblock 3dv: 3d dynamic voxel for action recognition in depth video.
\newblock In {\em Proceedings of the IEEE/CVF conference on computer vision and pattern recognition}, pages 511--520.

\bibitem[Wu et~al., 2023a]{wu2023LSM}
Wu, H., Hu, T., Luo, H., Wang, J., and Long, M. (2023a).
\newblock Solving high-dimensional pdes with latent spectral models.
\newblock In {\em International Conference on Machine Learning}.

\bibitem[Wu et~al., 2023b]{wu2023disentangling}
Wu, X., Lu, J., Yan, Z., and Zhang, G. (2023b).
\newblock Disentangling stochastic pde dynamics for unsupervised video prediction.
\newblock {\em IEEE Transactions on Neural Networks and Learning Systems}.

\bibitem[Xiao et~al., 2019]{xiao2019mvdi}
Xiao, Y., Chen, J., Wang, Y., Cao, Z., Zhou, J.~T., and Bai, X. (2019).
\newblock Action recognition for depth video using multi-view dynamic images.
\newblock {\em Information Sciences}, 480:287--304.

\bibitem[Yang et~al., 2023]{yang2023pde}
Yang, X., Shao, Y., Liu, S., Li, T.~H., and Li, G. (2023).
\newblock Pde-based progressive prediction framework for attribute compression of 3d point clouds.
\newblock In {\em Proceedings of the 31st ACM International Conference on Multimedia}, pages 9271--9281.

\bibitem[Zhong et~al., 2022]{Zhong2022kinet}
Zhong, J.-X., Zhou, K., Hu, Q., Wang, B., Trigoni, N., and Markham, A. (2022).
\newblock No pain, big gain: Classify dynamic point cloud sequences with static models by fitting feature-level space-time surfaces.
\newblock In {\em Proceedings of the IEEE/CVF Conference on Computer Vision and Pattern Recognition (CVPR)}, pages 8510--8520.

\end{thebibliography}
\bibliographystyle{apalike}

\newpage
\appendix
\onecolumn

\section{Details for the MHSA and MHCA layers}
\label{detailsMHSA}

The MHSA layer and MHCA layer share the same structure but with different inputs. The MHSA layer uses the same $F_t$ for query, key, and value generation. The MHCA layer uses the $F_s^{masked}$ for query generation, and the mapping output $F_{t\rightarrow s}$ for key and value generation.

\paragraph{MHSA}
Firstly, the $F_t$ is fed into a standard MHSA \citep{vaswani2017attention} module. We add the frame indexes as the position embedding. This process can be formulated as follows:
\begin{gather}
\label{eq:mhsa}
    F_t = \mathrm{PE}([1, 2, ..., T]) + F_t, \\
    H_m = \mathrm{Softmax}(\{F_tW_m^{Q}\} \times \{F_tW_m^{K}\}^{Transpose} / \sqrt{d}) \times F_tW_m^{V}, \\
    F_t = \mathrm{Concat}(H_1,\dots,H_m),
\end{gather}
where $\mathrm{PE}(\cdot)$ is the positional encoding function that embeds the frame index to high-dimension. $W_m^Q, W_m^K, W_m^V$ are learnable weights of the $m$th head for query, key, and value respectively. And $d$ is the number of feature channels. 

\paragraph{MHCA} 
After getting the mapping output $F_{t\rightarrow s}$ from our spectral method layer, we align the two different Banach spaces of $F_{t\rightarrow s}$ and $F_s$ by an MHCA layer. We execute this process with a learnable masked parameters set that is aligned with the $F_s$. This process can be formulated as follows:
\begin{gather}
\label{eq:mhca}
    H_m = \mathrm{Softmax}(\{F_s^{masked}W_m^{Q}\} \times \{F_{t\rightarrow s}W_m^{K}\}^{Transpose} / \sqrt{d}) \times F_{t\rightarrow s}W_m^{V}, \\
    \hat{F_s} = \mathrm{Concat}(H_1,\dots,H_m),
\end{gather}
the $F_s^{masked}$ is initialized with the same shape as $F_s$. We then match the predicted $\hat{F_s}$ and the ground truth $F_s$ as the supervision of the PDEs-solving.

\section{Visualization} 
\label{apx:vis}

We also report the feature response from PointNet++ \citep{qi2017pointnet++} as further comparison with our Motion PointNet. The observation reveals that the PointNet++ model exhibits a response to regions where geometric features are distinctly pronounced, such as the head, shoulders, and arms, irrespective of whether these areas constitute the primary focus of the action. As we can see, in our Motion PointNet, the main moving part of actions (\textit{e.g.} swinging arms in the golf swing) are highlighted, which is consistent with the proposed intuition. 

\begin{figure}[htbp]
\setlength{\abovecaptionskip}{.0cm}
\begin{center}
\includegraphics[width=\linewidth]{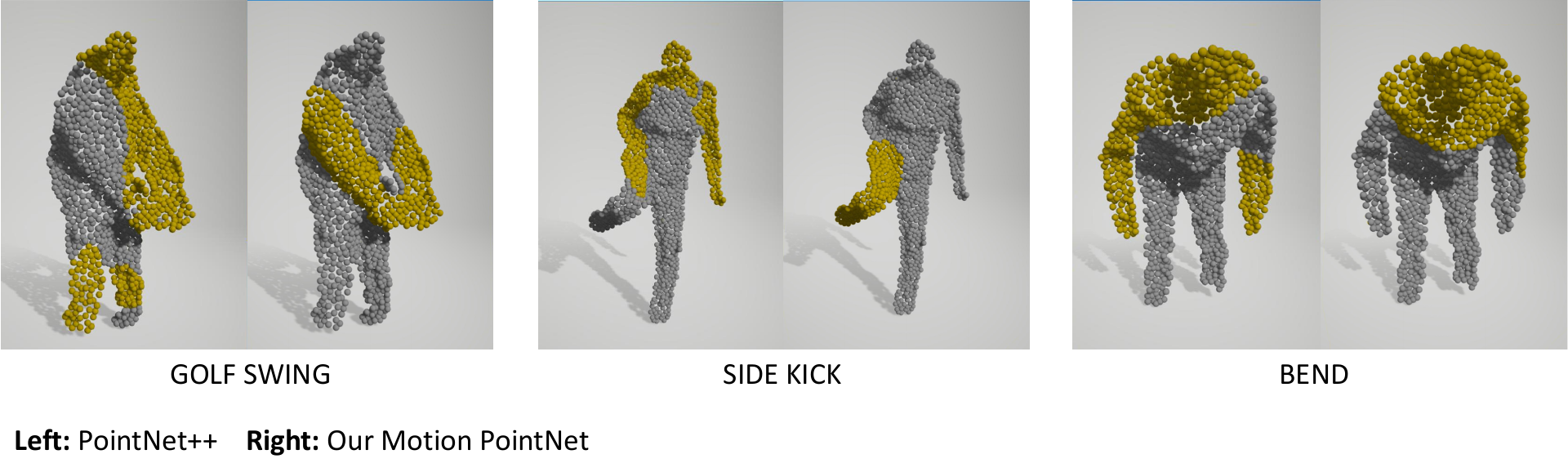}
\end{center}
\caption{Visualization comparison between PointNet++ and our Motion PointNet. High response points are marked in orange, which are selected based on the magnitude of the feature response. We choose binary representation for clarity in visualization.
}
\label{fig:feat}
\end{figure}

\begin{figure}[htbp]
\begin{center}
\includegraphics[width=\linewidth]{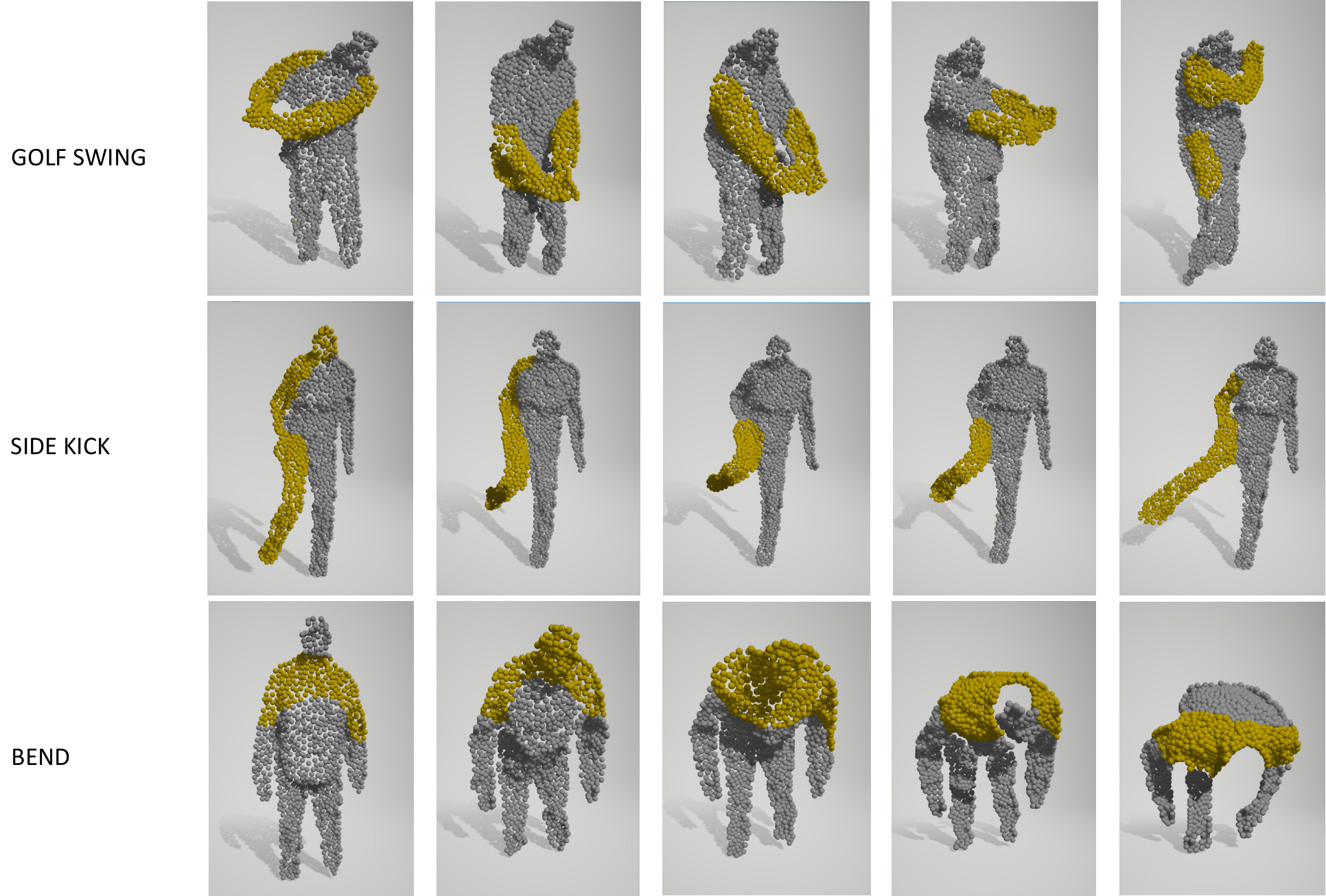}
\end{center}
\caption{Visualization of high feature response on MSRAction-3D dataset. High response points are marked in orange, which are selected based on the magnitude of the feature response. We choose binary representation for clarity in visualization.
}
\label{fig:feat}
\end{figure}

\end{document}